\documentclass{article}

\usepackage[nonatbib,final]{neurips_2021}

\usepackage[utf8]{inputenc} 
\usepackage[T1]{fontenc}    
\usepackage{booktabs}       
\usepackage{graphicx}
\usepackage[tableposition=top]{caption}
\usepackage{chngcntr}
\usepackage{hyperref}       
\usepackage{url}            

\title{A Data-Centric Behavioral Machine Learning Platform to Reduce Health Inequalities}

\author{
	Dexian Tang
	\quad
	Guillem Franc\`es
	\quad
	África Periáñez
	\\
	benshi.ai
	\\
	{\texttt{\{dexian,guillem,africa\}@benshi.ai}}
}
\begin{document}

\maketitle
\vspace{-0.1cm}
\begin{abstract}



Providing front-line health workers in low- and middle- income countries with recommendations and predictions to improve health outcomes can have a tremendous impact on reducing healthcare inequalities, for instance by helping to prevent the thousands of maternal and newborn deaths that occur every day%
. To that end, we are developing a data-centric machine learning platform that leverages the behavioral logs from a wide range of mobile health applications running in those countries. Here we describe the platform architecture, focusing on the details that help us to maximize the quality and organization of the data throughout the whole process, from the data ingestion with a data-science purposed software development kit to the data pipelines, feature engineering and model management.

\end{abstract}

\section{Introduction}
\label{sec:intro}
The rapid expansion of mobile health applications in low- and middle-income countries (LMICs), and the large volume of data generated by their users, has created unprecedented opportunities to use artificial intelligence to improve individual and population health~\cite{Hosny2019,Wahl2018}.
%
Those apps address some of the biggest challenges in global health, including midwifery, epidemiological diseases, diagnosis support and accessibility to medications.
Applying behavioral machine learning (ML) to digital health \cite{hrnjic2019machine} can improve the training of healthcare professionals and health service delivery, and benshi.ai is pursuing that goal by building a data-centric ML platform that ingests high-quality behavioral logs from a wide range of mobile health apps running in LMICs.




Our platform features a software development kit (SDK) to be integrated in the apps that ensures unified data schemas and ML-purposed labeling. This greatly facilitates feature engineering and ML processing. In order to maintain the quality of the information, the backend has a well-organized data structure and pipelines that focus on individual-user behavioral time-varying data and content consumption. 
 

The main challenges faced by ML platforms in LMICs are related to offline connections (most apps are able to run offline), and network latency due to scarce bandwidth and users turning off data to reduce cost. These issues affect data collection and users' ability to receive in-app incentives, hence they need to be solved. On the one hand, it is essential to batch event logs at the SDK level, uploading them to the platform only when there is a connection.
On the other, we also collect high-quality individual network information (e.g. the user’s connection speed and if an event log is generated while a user is online or offline), which is crucial when designing a system of incentives, in order to ensure that every user receives them on time.

\section{Data ingestion, structuring and validation}

To minimize latency and maximize quality, data is ingested by the platform directly
from the mobile devices of the end users. Organizations are provided with an SDK that allows their apps to send JSON-formatted app usage data to the platform servers. The purpose of the SDK is twofold: unifying behavioral data labeling and ensuring data format consistency. The former aspect, crucial when there are multiple data sources with their own vocabulary, is addressed by means of a set of built-in schema that facilitates data collection with the correct label (page views, purchases, etc.), while including essential information such as connection speed.
On the other hand, our SDK guarantees the consistency of timestamp or location formats,
thereby providing downstream data/ML pipelines with high-quality data that makes the ML lifecycle much smoother.

\subsection{Input data validation}
The ingested data is automatically validated against well-documented schemas that have been designed taking into account the needs of the data pipelines and ML models that will use this data \cite{breck2017ml}. The platform also provides an external API to retrieve the data ingested from the mobile apps, and also with a web interface to visualize and perform queries on this data.
This is a key mechanism not only to perform manual data validation and curation, but also to offer maximum transparency \cite{kostkova2015grand}.

\begin{figure*}
     \centering
     \includegraphics[width=.9\textwidth]{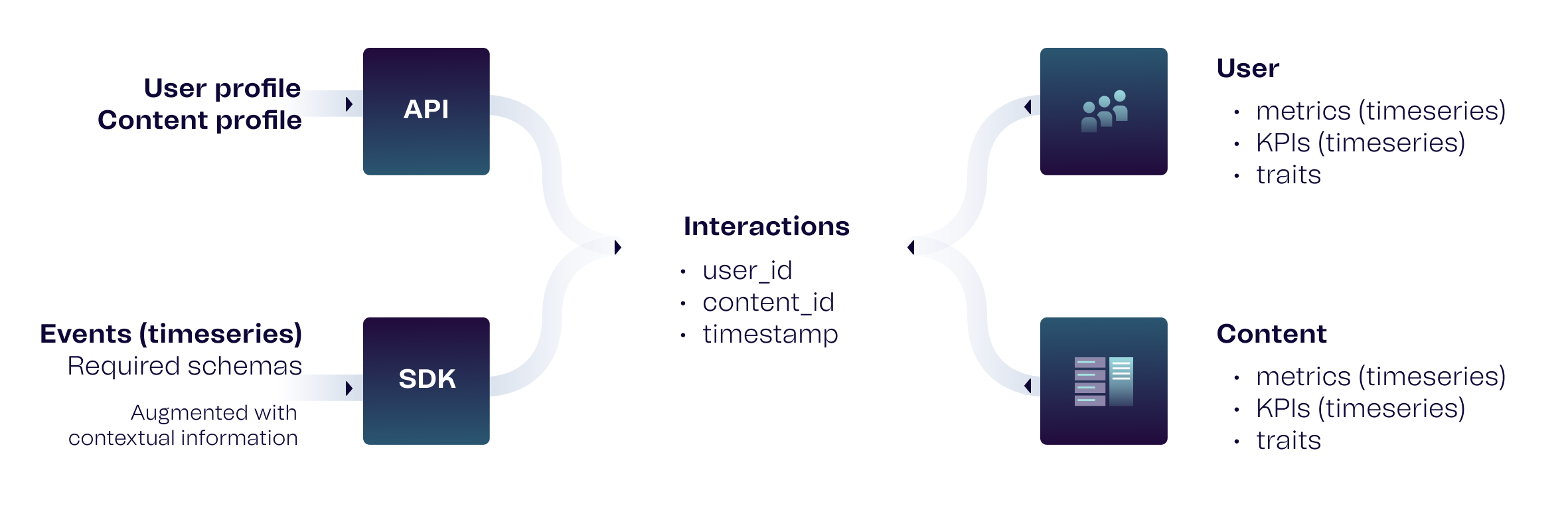}
     \caption{Data structure of time-varying behavioral data by user and content. On one side we organize the logs generated by individual users, on the other every interaction by each in-app content.}
     \label{fig:data-structure}
 \end{figure*}

\subsection{User and content data structure}
Fig.~\ref{fig:data-structure} shows an schema of organization of data by user and content. We consider three types of information: user data, content data (such as videos, learning modules, diagnosis...) and their interactions. The schema for user and content data are similar and include metrics, key performance indicators (KPIs) and traits. Metrics are individual-level time-series data, KPIs are non-individual time-series data aggregated from the metrics, and traits are dimensional data representing various attributes or statistics of a certain user or content. Finally, we capture the time-series, individual-level interactions between each user and content, which provides the foundation for personalized ML.

\section{Data pipelines and model management}

Once ingested, data runs through several data pipelines that 
prepare it for consumption by the ML models and other data analyses.
The pipelines are implemented using Python and Apache Spark~\cite{zaharia2010spark} on Databricks, on top of 
a Delta Lake storage layer~\cite{armbrust2020delta}. They include extensive pipeline-specific data
validation checks (running alongside the data transformation routines) to detect
data quality issues arising from incorrect source data or logical errors in the code \cite{breck2019data}.  
Delta Lake's \emph{time travel} facilitates the reproducibility of ML experiments
and gives data scientists full control over the various data snapshots.
MLflow~\cite{zaharia2018accelerating} is used for model management and deployability, and experiment tracking. 

\section{Conclusion}
We presented a data-centric behavioral ML platform to provide online recommendations within mobile health apps used by health workers and patients in LMICs.
Some challenges need to be addressed, including offline usage or network latency. We designed an SDK to appropriately record and label the logs that serve as input to the ML models, and also a user- and content-based data structure and data pipeline that organizes the time-varying data in a way that ensures its quality.

\section*{Acknowledgements} 
The authors wish to thank Javier Grande for their careful review of the manuscript and Marisa Asari for the design of the figure. This work was supported, in whole or in part, by the Bill \& Melinda Gates Foundation INV-022480. Under the grant conditions of the Foundation, a Creative Commons Attribution 4.0 Generic License has already been assigned to the Author Accepted Manuscript version that might arise from this submission.


\bibliographystyle{plain}
\bibliography{main}

\end{document}